\documentclass{article}

\usepackage[final]{nips_2018}
 
\usepackage[utf8]{inputenc} 
\usepackage[T1]{fontenc}    
\usepackage{hyperref}       
\usepackage{url}            
\usepackage{booktabs}       
\usepackage{amsfonts}       
\usepackage{nicefrac}       
\usepackage{microtype}      

\usepackage{multirow}
  
\usepackage[utf8]{inputenc} 
\usepackage[T1]{fontenc}    
\usepackage{hyperref}       
\usepackage{url}            
\usepackage{booktabs}       
\usepackage{amsfonts}       
\usepackage{nicefrac}       
\usepackage{microtype}      
\usepackage{graphicx}
\usepackage{tabularx}
\usepackage{amssymb,amsmath}
\usepackage{float}

\newcommand\clearrow{\global\let\rowmac\relax}
\clearrow

\title{Effective, Fast, and Memory-Efficient Compressed Multi-function Convolutional Neural Networks for More Accurate Medical Image Classification}

\begin{document}
\author {Luna M. Zhang}
\maketitle

\begin{abstract}
Convolutional Neural Networks (CNNs) usually use the same activation function, such as RELU, for all convolutional layers. There are performance limitations of just using RELU. In order to achieve better classification performance, reduce training and testing times, and reduce power consumption and memory usage, a new "Compressed Multi-function CNN" is developed. Google's Inception-V4, for example, is a very deep CNN that consists of 4 Inception-A blocks, 7 Inception-B blocks, and 3 Inception-C blocks. RELU is used for all convolutional layers. A new "Compressed Multi-function Inception-V4" ($CMI$) that can use different activation functions is created with $k$ Inception-A blocks, $m$ Inception-B blocks, and $n$ Inception-C blocks where $k\in\{1, 2, 3, 4\}$, $m\in\{1, 2, 3, 4, 5, 6, 7\}$, $n\in\{1, 2, 3\}$, and $(k+m+n)<14$. 
For performance analysis, a dataset for classifying brain MRI images into one of the four stages of Alzheimer's disease is used to compare three $CMI$ architectures with Inception-V4 in terms of F1-score, training and testing times (related to power consumption), and memory usage (model size). Overall, simulations show that the new $CMI$ models can outperform both the commonly used Inception-V4 and Inception-V4 using different activation functions. In the future, other "Compressed Multi-function CNNs", such as "Compressed Multi-function ResNets and DenseNets" that have a reduced number of convolutional blocks using different activation functions, will be developed to further increase classification accuracy, reduce training and testing times, reduce computational power, and reduce memory usage (model size) for building more effective healthcare systems, such as implementing accurate and convenient disease diagnosis systems on mobile devices that have limited battery power and memory.
\end{abstract}

\section{Introduction}

In recent years, deep learning techniques have been effectively used in various applications in computer vision, healthcare, etc. [1-22]. For example, the new 28nm Two-Dimensional Convolutional Neural Network (CNN)-DSA accelerator with an ultra power-efficient performance of 9.3 TOPS/Watt was implemented for low-end mobile and embedded platforms and MCUs (Microcontroller Units) [16]. Since DenseNets require large GPU memory, new methods were developed to reduce the memory consumption for training them [17]. Currently, it is especially important to build more effective, power-efficient, and memory-efficient CNNs for applications in the areas of internet of things (IoT), big data mining, green computing, mobile computing, etc. In particular, deep learning is important and useful in the evolution of mobile applications [18]. For instance, DeepVoice, a voiceprint-based Parkinsons disease identification software system, simultaneously integrates deep learning and mobile health [19]. Also, the CNN model (MobileNets) running on mobile devices can detect diabetic retinopathy with 73.3\% accuracy [20]. A personal and convenient diagnosis system of lung cancer is created using deep learning and sensor systems to achieve 90\% accuracy [21]. The adaptive single and multiple modality data compression methods using deep learning are used to build scalable and energy-efficient mobile health systems [22]. It is important to develop more effective, faster, and more energy-efficient deep learning systems for applications in mobile health informatics. 

Traditional CNNs usually use the same activation function (typically the rectified linear unit or RELU) for all convolutional layers. For example, Google’s very deep Inception-V4 network uses RELU [15]. However, traditional CNNs using RELU may not always be optimal in terms of model classification performance. Therefore, we design a new effective, fast, memory-efficient, and compressed CNN architecture ("Compressed Multi-function Convolutional Neural Networks) by using different activation functions and reducing the number of activation functions used. For simulations, Google's Inception-V4 architecture is used and modified to create a "Compressed Multi-function Inception-V4” ($CMI$) architecture. 

\section{Multi-function Convolutional Neural Networks}

CNNs mainly consist of convolutional layers, activation layers, pooling layers, and fully connected layers. For the convolutional layers, the purpose of convolution is to extract features from the input image. A filter slides over the input image (convolution operation) to produce an output image that is commonly called a feature map of
convolved features. There can be many convolutional layers.

An activation layer (AL) is defined as a layer of neurons where each neuron uses an activation function selected from a set of different activation functions. An AL transforms $m$ feature maps to new $m$ feature maps. The convolved feature $\theta^n$ is transformed to a new feature, which is $f(\theta^n)$, by an activation function $f$. Different activation functions can be used for different individual convolved features and different feature sub-maps. For the activation layers, a typical CNN uses a RELU layer right after each convolutional layer (CONV) and before a pooling layer (POOL) to generate $m$ new non-linear feature maps from the $m$ feature maps. For example, a typical CNN architecture is: INPUT $\rightarrow$ [[CONV $\rightarrow$ RELU] * M $\rightarrow$ POOL] * N $\rightarrow$ FCL $\rightarrow$ OUTPUT where FCL = fully connected layers. Let * indicate repetition and M $\geqslant$ 1 and N $\geqslant$ 1. The same activation function (RELU) is used to map all convolved features to new features. However, different activation functions can be used to build a Multi-function CNN (MCNN) to achieve better classification accuracy.

Let a convolutional block (CB) be a CONV followed by an AL. For example, a MCNN's CB is [CONV $\rightarrow$ AL(RELU, SIG)] where SIG = sigmoid function. The traditional CNN's CB is [CONV $\rightarrow$ REL], which is equivalent to the newly defined notation: [CONV $\rightarrow$ AL(RELU)]. AL(RELU) means that all neurons on the AL use RELU.

For example, Inception-V4 can be transformed to a Multi-function Inception-V4 (MI). Inception-V4 uses 4 Inception-A blocks, 7 Inception-B blocks, and 3 Inception-C blocks [15]. Inception-V4 consists of an Inception-Stem (11 CBs), Inception-A (7 CBs), Reduction-A (4 CBs), Inception-B (10 CBs), Reduction-B (6 CBs), and Inception-C (10 CBs).  The Inception-V4 with RELU has 149 CBs followed by RELU since 11 + 4$\times$7 + 4 + 7$\times$10 + 6 + 3$\times$10 = 149. A $MI$ also has 149 CBs but uses a variety of activation functions.

\section{Compressed Multi-function Convolutional Neural Networks}

CNNs using different activation functions have been shown to be able to achieve better performance than traditional CNNs that typically only use RELU. CNNs usually have a large number of layers and neurons, but their architectures can be reduced in size while still maintaining comparable, the same or even better performance. More specifically, the goal is to reduce the total of number of convolutional blocks and the total of number of activation functions in the whole deep neural network. 
Such a compressed architecture has faster training and testing times, less power usage, and less memory usage. Let this new architecture be called "Compressed Multi-function Convolutional Neural Networks." For Inception-V4, let it be called "Compressed Multi-function Inception-V4" ($CMI$). If there is no reduction in any of the layers, then let it be called "Multi-function Inception-V4" ($MI$).

To improve classification performance, reduce training and testing times, reduce power usage, reduce memory usage (model size), the new $CMI$ uses $k$ Inception-A blocks, $m$ Inception-B blocks, and $n$ Inception-C blocks where $k\in\{1, 2, 3, 4\}$, $m\in\{1, 2, 3, 4, 5, 6, 7\}$, $n\in\{1, 2, 3\}$, and $(k+m+n)<14$. For instance, a $CMI$ using 1 Inception-A block, 2 Inception-B blocks, and 1 Inception-C block has 58 CBs (58 activation functions) since 11 (one Inception-Stem) + 7 (one Inception-A) + 4 (one Reduction-A) + 2$\times$10 (2 Inception-Bs) + 6 (one Reduction-B) + 10 (one Inception-C) = 58. The $CMI$ using 58 CBs can run faster (use less power) and has a smaller model size (129MB) than the original Inception-V4 using 149 CBs with RELU, which has a larger model size of 323MB. The goal is to find a $CMI$ model with better performance, faster training and prediction times, less power usage, and less memory usage (model size) than the original Inception-V4.

\section{Experimental Results}

For simulations, Google's popular Inception-V4 architecture is used as the base framework. For testing purposes, the numbers of Inception-A blocks, Inception-B blocks, and Inception-C blocks are reduced. There are no changes to the inception stem, reduction A, and reduction B layers. The number of convolutional blocks for Inception-A, Inception-B, and Inception-C are unchanged too.

Let "$CMI_{i}$" and "$CI_{i}$" mean that a $CMI$ and a compressed Inception-V4 with RELU have $i$ Inception-A block(s), $i+1$ Inception-B block(s), and $i$ Inception-C block(s) for $i$ = 1, 2, and 3. This particular structure is just chosen to build several different smaller architectures for simulation purposes. 
For $CMI_{1}$, the number of activation functions is 58 as shown in the previous section. $CMI_{2}$ has 85 activation functions since 11 + 2$\times$7 + 4 + 3$\times$10 + 6 + 2$\times$10 = 85. $CMI_{3}$ has 112 since 11 + 3$\times$7 + 4 + 4$\times$10 + 6 + 3$\times$10 = 112. The model sizes of $CMI_{1}$, $CMI_{2}$, and $CMI_{3}$ are 129MB, 190MB, and 252MB, respectively. 
  
In the following tables, "$MI$" and “$I$” mean that a $MI$ and the original Inception-V4 with RELU have 4 Inception-A, 7 Inception-B, and 3 Inception-C blocks. The model size of $MI$ is 323MB.
Stratified 3-fold cross validation was used to evaluate and compare the three $CMI$ models, the $MI$ model, the three compressed Inception-V4 models with RELU, and the original Inception-V4 using multi-class classification metrics (i.e. training F1-score ($F1_{train}$), validation F1-scores ($F1_{valid}$), training times ($T_{train}$) in seconds, and classification testing times ($T_{test}$) in seconds. An activation function set \{RELU, SIG, TANH, ELU\} was used to build all of the multi-function models. Each activation function is randomly chosen from this set for each convolutional block. 
A dataset of 436 brain MRI images (cross-sectional collection of 416 subjects aged 18 to 96 and with extra data for 20 subjects), pre-processed and ready to be used, is used for performance analysis [23]. This research work uses all brain MRI images for a 4-class classification problem to determine the Alzheimer's Disease stage (non-demented, very mild dementia, mild dementia, or moderate dementia) of a person [23][24]. For each architecture ($CMI_{1}$, $CMI_{1}$, $CMI_{1}$, or $MI$), 10 random $CMI$ models and 10 random $MI$ models are created and tested. 

The highest cross-validation F1-score for each architecture is shown in Table 1 (100 training epochs). Table 1 shows that the $CMI_{3}$ model is the best. In addition, the three best $CMI$ models and one $MI$ model performed better than $CI_{1}$, $CI_{2}$, and $CI_{3}$ and the original Google’s Inception-v4 using RELU (i.e., $I$). $CMI_{2}$ and $CMI_{3}$ were better than $MI$.

\begin{table}[h!]
  \caption{Comparing the Best $CMI$ Models and $MI$ Model - 100 Training Epochs}
  \label{table1}
  \centering
  
\begin{tabular}{ c c c c c c c c c}
    \toprule
    Model: & $CMI_{1}$ & $CI_{1}$ & $CMI_{2}$ & $CI_{2}$ & $CMI_{3}$ & $CI_{3}$ & $MI$ & $I$ \\
 \midrule
    $F1_{train}$ & 0.77 & 0.75 & 0.83 & 0.72 & 0.84 & 0.74 & 0.83 & 0.73 \\
    $F1_{valid}$ & 0.77 & 0.76 & 0.82 & 0.72 & 0.83 & 0.73 & 0.80 & 0.74 \\
    $T_{train}$ (s) & 1871 & 1839 & 2563 & 2539 & 3217 & 3158 & 4212 & 4048 \\
    $T_{test}$ (s) & 1.30 & 1.21 & 1.57 & 1.49 & 1.83 & 1.78 & 2.40 & 2.30 \\
    \bottomrule
  \end{tabular}
\end{table}

Average performance results for 10 $CMI_{1}$ models, 10 $CMI_{2}$ models, 10 $CMI_{3}$ models, and 10 $MI$ models are shown in Table 2 (100 training epochs). $CMI_{3}$ model (252MB) can perform better, run faster (use less power) and use less memory than the $MI$ model (323MB).

\begin{table}[h!]
  \caption{Average Performance of 30 $CMI$ Models and 10 $MI$ Models - 100 Training Epochs}
  \label{table3}
  \centering
  \begin{tabular}{c c c c c}
    \toprule
    Model: & $CMI_{1}$ & $CMI_{2}$ & $CMI_{3}$  & $MI$  \\
 \midrule
    Avg. $F1_{train}$ & 0.753 & 0.780 & 0.786 & 0.783 \\
    Avg. $F1_{valid}$ & 0.747 & 0.773 & 0.780 & 0.771 \\
    Avg. $T_{train}$ (s) & 1867 & 2560 & 3237 & 4144 \\
    Avg. $T_{test}$ (s) & 1.31 & 1.55 & 1.86 & 2.36 \\
    \bottomrule
  \end{tabular}
\end{table}

The highest cross-validation F1-score for each architecture is shown in Table 3 (120 training epochs). Table 3 shows that the $CMI_{3}$ model is the best. In addition, the three best $CMI$ models and one $MI$ model always performed better than both the three compressed Inception-V4 models with RELU ($CI_{1}$, $CI_{2}$ and $CI_{3}$) and the original Google’s Inception-v4 using RELU.

\begin{table}[h!]
  \caption{Comparing the Best $CMI$ Models and $MI$ Model - 120 Training Epochs}
  \label{table1}
  \centering
  \begin{tabular}{ c c c c c c c c c}
    \toprule
    Model: & $CMI_{1}$ & $CI_{1}$ & $CMI_{2}$ & $CI_{2}$ & $CMI_{3}$ & $CI_{3}$ & $MI$ & $I$ \\
 \midrule
    $F1_{train}$ & 0.78 & 0.68 & 0.85 & 0.67 & 0.86 & 0.70 & 0.84 & 0.70 \\
    $F1_{valid}$ & 0.76 & 0.67 & 0.82 & 0.66 & 0.86 & 0.68 & 0.83 & 0.68 \\
    $T_{train}$ (s) & 2229 & 2202 & 3081 & 3021 & 3978 & 3816 & 5082 & 4815 \\
    $T_{test}$ (s) & 1.24 & 1.20 & 1.54 & 1.49 & 1.88 & 1.80 & 2.41 & 2.31 \\
    \bottomrule
  \end{tabular}
\end{table}

Average performance results for 10 $CMI_{1}$ models, 10 $CMI_{2}$ models, 10 $CMI_{3}$ models, and 10 $MI$ models are shown in Table 4 (120 training epochs). $CMI_{3}$ model (252MB) can perform better, run faster (use less power) and use less memory than the $MI$ model (323MB). 

\begin{table}[h!]
  \caption{Average Performance of 30 CMI Models and 10 $MI$ Models - 120 Training Epochs}
  \label{table3}
  \centering
  \begin{tabular}{c c c c c}
    \toprule
    Model: & $CMI_{1}$ & $CMI_{2}$ & $CMI_{3}$  & $MI$  \\
 \midrule
    Avg. $F1_{train}$ & 0.75 & 0.78 & 0.80 & 0.80 \\
    Avg. $F1_{valid}$ & 0.74 & 0.76 & 0.79 & 0.78 \\
    Avg. $T_{train}$ (s) & 2242 & 3078 & 3883 & 4964 \\
    Avg. $T_{test}$ (s) & 1.28 & 1.55 & 1.86 & 2.35 \\
    \bottomrule
  \end{tabular}
\end{table}

\section{Conclusions}

Simulation results show that $CMI$ can achieve better performance, shorter training and testing times (i.e., less power consumption), and less memory usage (model size) than $MI$, $CI_{1}$, $CI_{2}$, $CI_{3}$, and $I$. Thus, it is feasible to find better compressed Inception-V4 models using different activation functions than the traditional Inception-V4 model. Importantly, compressed MCNNs using a smaller number of convolutional blocks with different activation functions can be useful for applications in healthcare such as improving the accuracy and speed of disease diagnosis. Also, with faster training and classification times, computer systems and mobile devices running compressed MCNNs can save power, which would increase battery life, which is especially important for mobile health-based systems. In addition, the compressed MCNN that uses less memory is more memory-efficient than an original MCNN. These excellent properties of the compressed MCNN would be very useful for various intelligent health-based systems on mobile devices. 

\section{Future Works}

In the future, more $CMI$ models will be created and tested by reducing the number of convolutional blocks. Also, better automatic optimization algorithms will be developed to efficiently find the most effective, power-efficient, and memory-efficient $CMI$ models for different healthcare applications. Other compressed multi-function deep neural networks, such as compressed multi-function ResNets and compressed multi-function DenseNets with a reduced number of convolutional blocks using different activation functions, will be developed to increase classification accuracy, reduce training and testing times, reduce computational power, and reduce memory usage (model size) for effective healthcare application systems such as disease diagnosis systems on mobile devices with limited power on battery and small memory. Other medical datasets will be used to further test the new models. For effective and efficient medical applications, the optimized power-memory efficient compressed MCNN models will be pre-trained and embedded into mobile devices.

\pagebreak

\section*{References}

[1] LeCun, Y., Bengio, Y.\ \& Hinton, G.E. (2015) Deep learning. Nature 521, pp.\ 436--444.

[2] Krizhevsky, A., Sutskever, I.\ \& Hinton, G.E.\ (2012) Imagenet classification with deep convolutional neural networks. In {\it Advances in Neural Information Processing Systems 25}, pp.\ 1097--1105. Cambridge, MA: MIT Press.

[3] He, K., Zhang, X., Ren, S.\ \& Sun, J. \ (2016) Deep Residual Learning for Image Recognition. In Proceedings of the 2016 IEEE Conference on Computer Vision and Pattern Recognition (CVPR), pp.\ 770--778.

[4] Esteva, A., Kuprel, B., Novoa, R.A., Ko, J., Swetter, S.M., Blau, H.M.\ \& Thrun, S.\ (2017) Dermatologist-level classification of skin cancer with deep neural networks. {\it Nature} {\bf 542}(7639):115--118.

[5] Nugraha, B.T, Su, S.-F.\ \& Fahmizal, F.\ (2017) Towards self-driving car using convolutional neural network and road lane detector. In Proceedings of the 2nd International Conference on Automation, Cognitive Science, Optics, Micro Electro-Mechanical System, and Information Technology (ICACOMIT), pp.\ 65--69. 

[6] Silver, D., Huang, A., Maddison, C.J., Guez, A., Sifre, L., Driessche, G.V.D., Schrittwieser, J., Antonoglou, I., Panneershelvam, V., Lanctot, M., Dieleman, S., Grewe, D., Nham, J., Kalchbrenner, N., Sutskever, I., Lillicrap, T., Leach, M., Kavukcuoglu, K., Graepel, T.\ \& Hassabis, D.\ (2016) Mastering the game of Go with deep neural networks and tree search. Nature (529), pp 484--503. 

[7] Fukushima, K.\ (1979) Neural network model for a mechanism of pattern recognition
unaffected by shift in position-Neocognitron. {\it Transactions of the IECE} {\bf J62-A}(10):658--665.
 
[8] LeCun, Y., Bottou, L., Bengio, Y.\ \& Haffner, P.\ (1998) Gradient-based learning applied to document recognition. {\it Proc. IEEE} {\bf 86}(11):2278--2324.

[9] {\it Large Scale Visual Recognition Challenge 2012 (ILSVRC2012)} (2012)  [Online.] Available: http://www.image-net.org/challenges/LSVRC/2012/results.html.

[10] Szegedy, C., Liu, W., Jia, Y., Sermanet, P., Reed S., Anguelov D., Erhan, D., Vanhoucke, V.\ \& Rabinovich, A.\ (2015) Going Deeper with Convolutions. In Proceedings of 2015 IEEE Conference on Computer Vision and Pattern Recognition (CVPR), pp.\ 1--9. 

[11] {\it Large Scale Visual Recognition Challenge 2014 (ILSVRC2014)} (2014)  [Online.] Available: http://www.image-net.org/challenges/LSVRC/2014/.

[12] {\it Large Scale Visual Recognition Challenge 2015 (ILSVRC2015)} (2015) [Online.] Available: http://www.image-net.org/challenges/LSVRC/2015/index.

[13] He K.\ (2016) 
Deep Residual Networks - 
Deep Learning Gets Way Deeper. 
[Online.] Available: 
https://icml.cc/2016/tutorials/icml2016\textunderscore tutorial\textunderscore deep\textunderscore residual\textunderscore networks\textunderscore kaiminghe.pdf.

[14] {\it COCO 2015 Object Detection Task} (2015) [Online.] Available: http://cocodataset.org/\#detection-2015.

[15] Szegedy, C., Ioffe, S., Vanhoucke, V.\ \& Alemi, A.\ (2017) Inception-v4, Inception-ResNet and the Impact of Residual Connections on Learning. In Proceedings of the Thirty-First AAAI Conference on Artificial Intelligence (AAAI-17), pp.\ 4278--4284.

[16] Pleiss G., Chen D., Huang G., Li T., Maaten L. v. d.\ \& Weinberger K. Q.\ (2017) Memory-Efficient Implementation of DenseNets. [Online.] Available: https://arxiv.org/abs/1707.06990.

[17] Sun B., Yang L., Dong P., Zhang W., Dong J.\ \& Young C.\ (2018) Ultra Power-Efficient CNN Domain Specific Accelerator with 9.3TOPS/Watt for Mobile and Embedded Applications. [Online.] Available: https://arxiv.org/abs/1805.00361.

[18] Wang, J., Yu, P., Sun, L., Bao, W.\ \& Zhu, X.\ (2018) Deep Learning Towards Mobile Applications. [Online.] Available:  https://arxiv.org/pdf/1809.03559.pdf.

[19] Zhang H., Wang A., Li D.\ \& Xu W.\ (2018) DeepVoice: A voiceprint-based mobile health framework for Parkinson's disease identification. 2018 IEEE EMBS International Conference on Biomedical \& Health Informatics (BHI), pp.\ 214--217.

[20] Suriyal S., Druzgalski C.\ \& Gautam K.\ (2018) Mobile assisted diabetic retinopathy detection using deep neural network. 2018 Global Medical Engineering Physics Exchanges/Pan American Health Care Exchanges (GMEPE/PAHCE), pp.\ 1--4.

[21] Shimizu R., Yanagawa S., Monde Y., Yamagishi H., Hamada M., Shimizu T.\ \& Kuroda T.\ (2018) Deep learning application trial to lung cancer diagnosis for medical sensor systems. 2016 International SoC Design Conference (ISOCC), pp.\ 191--192. 

[22] Said A. B., Al-Sa’D M. F., Tlili M., Abdellatif A. A., Mohamed A., Elfouly T., Harras K.\ \& O’Connor M. D.\ (2018) A Deep Learning Approach for Vital Signs Compression and Energy Efficient Delivery in mhealth Systems. IEEE Access, volume: 6, pp.\ 33727--33739.

[23] {\it OASIS Brains Datasets}. [Online]. Available:  http://www.oasis-brains.org/\#data.

[24] Marcus, D.S., Wang, T.H., Parker, J., Csernansky, J.G., Morris,
J.C.\ \& Buckner, R.L.\ (2007) Open access series of imaging studies (oasis):
cross-sectional MRI data in young, middle aged, nondemented, and
demented older adults. Journal of cognitive neuroscience 19(9),
1498--1507.

\end{document}